\documentclass[10pt,twocolumn,letterpaper]{article}

\usepackage{cvpr}
\usepackage{times}
\usepackage{epsfig}
\usepackage{graphicx}
\usepackage{amsmath}
\usepackage{mathtools}
\usepackage{amssymb}
\usepackage{array}
\usepackage{float}
\usepackage{color}
\usepackage{breqn}

\newenvironment{commentEnv}[1]{\color{red}\textbf{#1:}\ }{}

\makeatletter
\def\blfootnote{\gdef\@thefnmark{}\@footnotetext}
\makeatother

\newcolumntype{?}{!{\vrule width 1pt}}
\usepackage[pagebackref=true,breaklinks=true,letterpaper=true,colorlinks,bookmarks=false]{hyperref}

\cvprfinalcopy % *** Uncomment this line for the final submission

\def\cvprPaperID{3817} % *** Enter the CVPR Paper ID here

\ifcvprfinal\fi
\begin{document}

%%%%%%%%% TITLE
\title{Features for Multi-Target Multi-Camera Tracking and Re-Identification}

\author{Ergys Ristani \quad Carlo Tomasi\\
Duke University\\
Durham, NC, USA\\
{\tt\small \{ristani, tomasi\}@cs.duke.edu}
%\and
}

\maketitle

\begin{abstract}
Multi-Target Multi-Camera Tracking (MTMCT)\blfootnote{This material is based upon work supported by the National Science Foundation under Grants No. IIS-1420894 and CCF-1513816.} tracks many people through video taken from several cameras. Person Re-Identification (Re-ID) retrieves from a gallery images of people similar to a person query image. We learn good features for both MTMCT and Re-ID with a convolutional neural network. Our contributions include an adaptive weighted triplet loss for training and a new technique for hard-identity mining. Our method outperforms the state of the art both on the DukeMTMC benchmarks for tracking, and  on the Market-1501 and DukeMTMC-ReID benchmarks for Re-ID. We examine the correlation between good Re-ID and good MTMCT scores, and perform ablation studies to elucidate the contributions of the main components of our system. Code is available\footnote{\url{http://vision.cs.duke.edu/DukeMTMC/}}.

%At test time, we use these features to compute correlations for an existing state-of-the-art data association method that groups observations into identities. The input to our processing pipeline comes from an existing state-of-the-art person detector. To scale with the large number of detections we reason hierarchically about co-identity over temporal intervals that increase in duration. 

%Multi-Target Multi-Camera Tracking (MTMCT) tracks a multitude of people through video taken from several cameras. We propose a new combination of a weighted triplet loss and hard negative mining for training a CNN. The features we learn outperform the competition in both MTMCT and person re-identification (ReID). We also found that improving ReID performance improves MTMCT performance only up to a point

\end{abstract}

\section{Introduction}

Multi-Target Multi-Camera Tracking (MTMCT) aims to determine the position of every person at all times from video streams taken by multiple cameras. The resulting multi-camera trajectories enable applications including visual surveillance, suspicious activity and anomaly detection, sport player tracking, and crowd behavior analysis. In recent years, the number of cameras has increased dramatically in airports, train stations, and shopping centers, so it has become necessary to automate MTMC tracking.

MTMCT is a notoriously difficult problem: Cameras are often placed far apart to reduce costs, and their fields of view do not always overlap. This results in extended periods of occlusion and large changes of viewpoint and illumination across different fields of view. In addition, the number of people is typically not known in advance, and the amount of data to process is enormous.

\begin{figure}[htb]
	\centering
	\includegraphics[scale = 0.54]{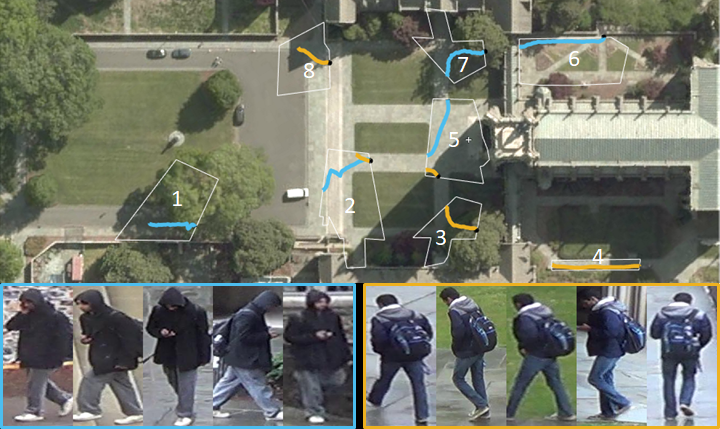}
	
	\label{fig:example}
	
	\caption{Two example multi-camera results from our tracker on the DukeMTMC dataset.}
\end{figure}

Person re-identification (Re-ID) is closely related to MTMCT: Given a snapshot of a person (the query), a Re-ID system retrieves from a database a list of other snapshots of people, usually taken from different cameras and at different times, and ranks them by decreasing similarity to the query. The intent is that any snapshots in the database that are \emph{co-identical} with (that is, depict the same person as) the person in the query are ranked highly.

MTMCT and Re-ID differ subtly but fundamentally, because Re-ID \emph{ranks} distances to a query while MTMCT \emph{classifies} a pair of images as being co-identical or not, and their performance is consequently measured by different metrics: ranking performance for Re-ID, classification error rates for MTMCT. This difference would seem to suggest that appearance features used for the two problems must be learned with different loss functions. Ideally, the Re-ID loss ought to ensure that \emph{for any query $a$} the largest distance between $a$ and a feature that is co-identical to it is smaller than the smallest distance between $a$ and a feature that is not co-identical to it. This would guarantee correct feature ranking for any given query. In contrast, the MTMCT loss ought to ensure that the largest distance between \emph{any two} co-identical features is smaller that the smallest distance between \emph{any two} non co-identical features, to guarantee a margin between within-identity and between-identity distances.

With these criteria, zero MTMCT loss would imply zero Re-ID loss, but not \textit{vice versa}. However, training with a loss of the MTMCT type is very expensive, because it would require using all pairs of features as input. More importantly, there would be a severe imbalance between the number of within-identity pairs and the much greater number of between-identity pairs. In this paper, we couple a triplet loss function of the Re-ID type with a training procedure based on hard-data mining and obtain high-performing features for both Re-ID and MTMCT. Our experiments also show that when tracking moderately crowded scenes, improving Re-ID rank accuracy beyond a certain point shows diminishing returns for MTMCT.

\begin{figure*}[t]
	\centering
	\includegraphics[scale = 0.5]{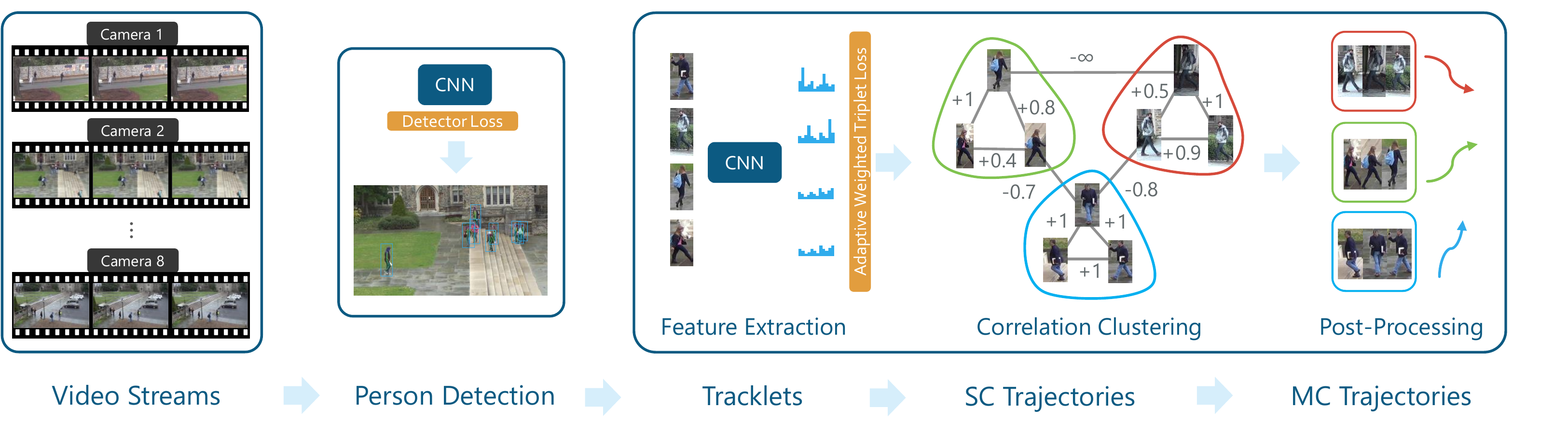}
	\caption{An illustration of our pipeline for Multi-Target Multi-Camera Tracking. Given video streams, a person detector extracts bounding box observations from video. For trajectory inference, a feature extractor extracts motion and appearance features from observations. These are in turn converted into correlations and labeled using correlation clustering optimization. Finally, post-processing interpolates missing detections and discards low confidence tracks. Multi-stage reasoning repeats trajectory inference for tracklets, single- and multi-camera trajectories. At train time the detector is trained independently, and the feature loss penalizes features that yield wrong correlations.}
	\label{fig:pipeline}
\end{figure*}

To use our features for MTMCT, we assemble a processing pipeline (Figure \ref{fig:pipeline}) that uses a state-of-the-art person detector and, at test time, a state-of-the art data association algorithm based on correlation clustering to group observations into identities. To reduce computational complexity, we also incorporate standard hierarchical reasoning and sliding temporal window techniques in our tracker. Some qualitative results from our method are shown in Figure~\ref{fig:example}.

We do \emph{not} include correlation clustering when training. Instead, we make the conjecture that high-quality appearance features lead to good clustering solutions, and only train the features. Our state-of-the art experimental results on the DukeMTMCT benchmark bear out this conjecture. %, and we also show state-of-the art performance on several Re-ID benchmarks.

In summary we make the following contributions:
\begin{itemize}
    \setlength\itemsep{0em}
    \item We propose an adaptive weighted triplet loss that, unlike fixed-weight variants, is both accurate and stable.
    \item We propose an inexpensive hard-identity mining scheme that helps learn better features.
    \item We provide new insights on the relation between tracking and ranking accuracy on existing benchmarks.
    \item We show experimentally that our features yield state-of-the-art results on both MTMCT and Re-ID tasks.
\end{itemize}

\section{Related Work}

We summarize work on different aspects of MTMCT.

\noindent\textbf{Person Detection.} MTMC trackers rely on person detection and some trackers assume that single-camera trajectories are given~\cite{bredereck_data_2012,cai_exploring_2014,chen_adaptive_2011,chen_multitarget_2015,chen_direction-based_2011,daliyot_framework_2013,das2014consistent,gilbert_tracking_2006,javed_modeling_2008,kuo_intercamera_2010,makris_bridging_2004,zhang_camera_2015}. The popular Deformable Parts Model detector~\cite{felzenszwalb2010object} was used as the public detector for MOTChallenge sequences~\cite{MOTChallenge2015, milan2016mot16, ristani2016performance} and in labeling Re-ID datasets~\cite{zheng2016mars, zheng2015scalable}. Since the MOT17 challenge, trackers have shown increased accuracy by utilizing detectors that rely on deep learning. These include Faster R-CNN~\cite{ren2015faster}, SSD~\cite{liu2016ssd}, KDNT~\cite{yu2016poi}, or pose-based detectors~\cite{cao2017realtime, insafutdinov2016deepercut}. We use OpenPose \cite{cao2017realtime} which has shown good performance.

\noindent\textbf{Data Association.}
Most existing formulations, with some exceptions~\cite{beyer2017towards, Milan:2017:AAAI_RNNTracking,  Milan:2014:CEM}, are special cases of the multidimensional assignment problem~\cite{collins2012multitarget}: Input detections are arranged in a graph whose edges encode similarity and whose nodes are then partitioned into identities. Formulations with polynomial time solutions consider evidence along paths of time-consecutive edges~\cite{BerclazFTF11, CaoCCZH15, Fleuret08a,Izadinia_ECCV12_MP2T,javed_modeling_2008,jiuqing_distributed_2013, pirsiavash2011globally,zhang2008global,zhang_tracking_2015} and some build on bipartite matching~\cite{brendel2011multiobject, cai_exploring_2014,chen_multitarget_2015,daliyot_framework_2013, kuo_intercamera_2010, shu2012part,wu2007detection}. Methods that use all pairwise terms, not only time-consecutive ones, are significantly more accurate but NP-hard~\cite{chari2015pairwise,collins2012multitarget,das2014consistent, dehghan2015gmmcp,kumar2014multiple,ristani2014tracking, shafique2005noniterative,tang2015subgraph,tang2016multi,tang2017multiple}. Unary terms are sometimes added for completeness~\cite{dehghan2015gmmcp, tang2015subgraph}. Higher order terms have also been used~\cite{butt2013multiple, wen2014multiple} but with sharply diminishing returns.  Identities can be optimized jointly~\cite{dehghan2015gmmcp} or iteratively~\cite{ZamirECCV12}. We choose correlation clustering~\cite{bansal2002correlation, ristani2016performance, ristani2014tracking} to trade off computational cost for simplicity of formulation and accuracy. This formulation considers evidence from all pairwise terms and optimizes identities jointly. An equivalent formulation is that of graph multicuts ~\cite{tang2016multi} which minimizes disagreement instead of maximizing agreement~\cite{demaine2006correlation}.

\noindent\textbf{Appearance.} Human appearance has been described by color ~\cite{cai_exploring_2014,chen_adaptive_2011,chen_multitarget_2015,chen_direction-based_2011,das2014consistent,gilbert_tracking_2006,javed_modeling_2008,jiuqing_distributed_2013, kuo_intercamera_2010, zhang_camera_2015, zhang_tracking_2015} % on foreground masks~\cite{chen_direction-based_2011,daliyot_framework_2013,das2014consistent,gilbert_tracking_2006,jiuqing_distributed_2013}
and texture descriptors~\cite{cai_exploring_2014,chen_multitarget_2015,daliyot_framework_2013,kuo_intercamera_2010,zhang_camera_2015,zhang_tracking_2015}. Lighting variations are addressed through color normalization~\cite{cai_exploring_2014}, exemplar-based approaches~\cite{chen_multitarget_2015}, or brightness transfer functions learned with~\cite{das2014consistent,javed_modeling_2008} or without supervision~\cite{chen_adaptive_2011,gilbert_tracking_2006, zhang_camera_2015, zhang_tracking_2015}.
Discriminative power is improved by \emph{saliency} information~\cite{martinel2014saliency,zhao2013unsupervised} or by \emph{learning} features specific to body parts~\cite{cai_exploring_2014,chen_multitarget_2015,chen_direction-based_2011,daliyot_framework_2013,das2014consistent,jiuqing_distributed_2013, kuo_intercamera_2010}, either in the image~\cite{BedagkarGala6130457,BedagkarGala20121908,ChengBMVC2568} or back-projected onto an articulated~\cite{learning_baltieri_2013,Shaogang14} or monolithic~\cite{mapping_baltieri_2015} 3D body model. The current state of the art in person re-identification relies on deep learning~\cite{zheng2016person,zheng2017unlabeled}, hard negative mining~\cite{zheng2017unlabeled}, data augmentation~\cite{barbosa2017looking,zhong2017random}, special purpose layers~\cite{SunZDW17} or branches~\cite{zheng2017pedestrian}, and specialized loss functions~\cite{hermans2017defense}. We use a residual network ~\cite{he2016deep} and similar techniques to learn good features for MTMCT and Re-ID.

\noindent\textbf{Multiple Cameras.} Spatial relations between cameras are either explicitly mapped in 3D~\cite{chen_adaptive_2011, zhang_camera_2015}, learned by tracking known identities~\cite{4407431, javed_modeling_2008,jiuqing_distributed_2013}, or obtained by comparing entry/exit rates across pairs of cameras~\cite{cai_exploring_2014, kuo_intercamera_2010,makris_bridging_2004}.
%, or discovered on-line~\cite{gilbert_tracking_2006,zhang_tracking_2015}.
Pre-processing methods may fuse data from partially overlapping views~\cite{zhang_tracking_2015}, while some systems rely on completely overlapping and unobstructed views~\cite{ayazoglu_dynamic_2011, BerclazFTF11,bredereck_data_2012,hamid_player_2010, kamal_information_2013}. People \textit{entry and exit points} may be explicitly modeled on the ground~\cite{cai_exploring_2014,chen_adaptive_2011,kuo_intercamera_2010,makris_bridging_2004} or image plane~\cite{gilbert_tracking_2006,jiuqing_distributed_2013}. 
%, and may handle changes over time~\cite{chen_adaptive_2011}.
\textit{Travel time} is also modeled, either parametrically~\cite{jiuqing_distributed_2013, zhang_camera_2015} or not~\cite{chen_adaptive_2011,gilbert_tracking_2006,javed_modeling_2008,kuo_intercamera_2010, makris_bridging_2004}. We use time constraints to rule out unlikely inter-camera associations. Similarly to~\cite{ristani2016performance} we decay correlations to zero as the time distance between observations increases. Correlation decay ensures that time-distant observations are associated if there is a chain of positively-correlated observations that connect them. The idea is similar to lifted multicuts~\cite{tang2017multiple}, although we employ no threshold or hard constraints.

\noindent\textbf{Learning to Track.} There have been several attempts to learn multi-target tracking data association in a supervised way, either through recurrent neural networks for end-to-end prediction of trajectories~\cite{Milan:2017:AAAI_RNNTracking} or by learning data association by back-propagating through a network-flow solution~\cite{DeepFlow}. These methods have been pushing in the right direction even though they haven't yet topped single-camera tracking benchmarks. In our method we learn features for correlations without measuring trajectory quality through combinatorial optimization. Our argument is that if correlations are good, even greedy association suffices. This idea has been shown to work for person detection~\cite{cao2017realtime}, and implicitly pursued in single-camera trackers~\cite{tang2016multi, tang2017multiple, yu2016poi} and Re-ID methods~\cite{hermans2017defense, zheng2016person, zheng2017pedestrian} that improve features to increase accuracy. Learning good correlations makes training simpler and less expensive, and we show that it achieves state-of-the-art performance for MTMCT.

\section{Method}

The input is a set of videos $V = \{V_1, \ldots, V_n\}$ from $n$ different cameras, and the ground truth is a set of multi-camera trajectories $T = \{ T_1, \ldots, T_{\ell}\}$. MTMCT could be cast as a supervised learning problem: Find the optimal parameters $\Theta^*$ of a function $f(\Theta, V)$ that estimates the true trajectories as well as possible:
\begin{equation}
\Theta^* = \arg\min_{\Theta} \mathcal{L}( f(\Theta,V), T)
\end{equation}  
where the loss function $\mathcal{L}$ could be derived from the multi-camera tracking accuracy measure IDF1~\cite{ristani2016performance}. 

However, end-to-end training would require back-propagating the loss through a combinatorial optimization layer that performs data association, and this is expensive~\cite{DeepFlow}. We avoid this complexity by noting that if the correlations were positive for co-identical pairs and negative for non co-identical pairs, then combinatorial optimization would be trivial. Thus, we aim to learn features that produce good correlations during training, while at test time we employ correlation clustering to maximize agreement between potentially erroneous correlations.

An additional source of difficulty during training is model depth, as weight updates can fail to propagate back to early layers responsible for person detection. If the network is monolithic and trained with a single loss, training becomes more difficult. We therefore separate detection and association as is customary in the literature (Figure \ref{fig:pipeline}). In the following we describe how we learn appearance features, and the different parts of the tracker.

%\begin{figure*}[t]
%	\centering
%	\includegraphics[scale = 0.5]{figures/hard_mining.pdf}
%	\caption{An illustration of our hard mining strategy. For each anchor identity, half of the $P-1$ identities in the batch are sampled from the hard identity pool, the other half from the random identity pool. Hard-negative identities are outlined in red and correct matches in green.}
%	\label{fig:hard_mining}
%\end{figure*}

\begin{figure*}[ht]
    \centering
    \begin{minipage}{.4\textwidth}
        \centering
        \includegraphics[height=0.2\textheight]{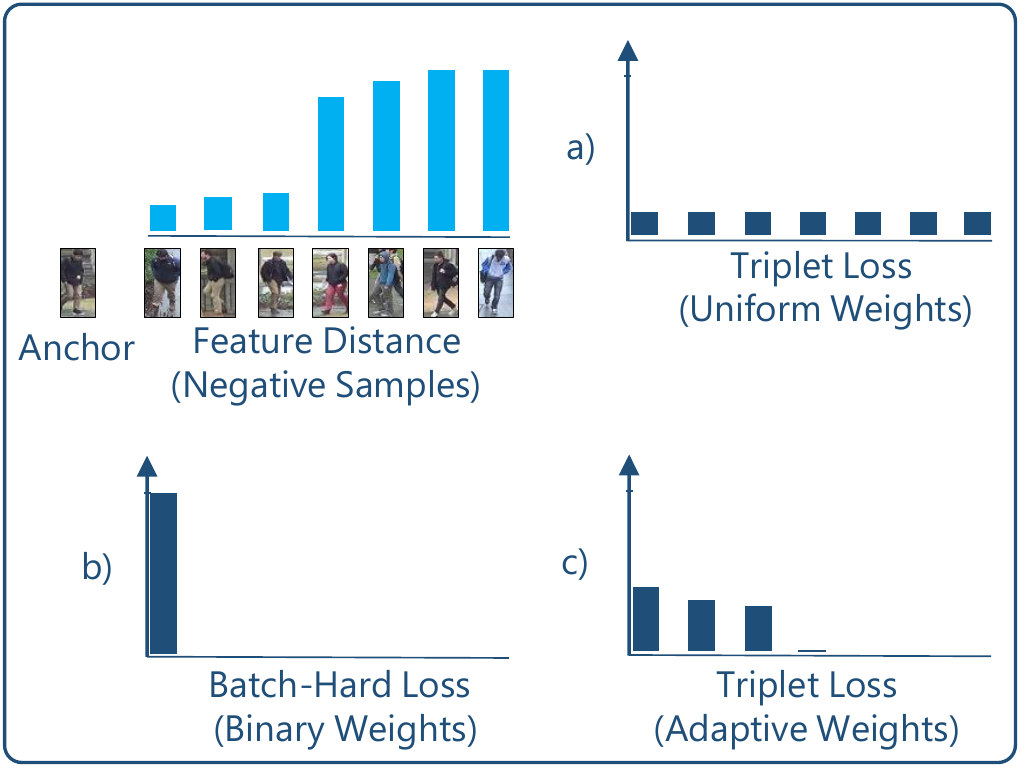}
        \caption{Triplet loss weighing schemes.}
        \label{fig:weighted_loss}
    \end{minipage}%
    \begin{minipage}{0.6\textwidth}
        \centering
    	\includegraphics[scale = 0.45]{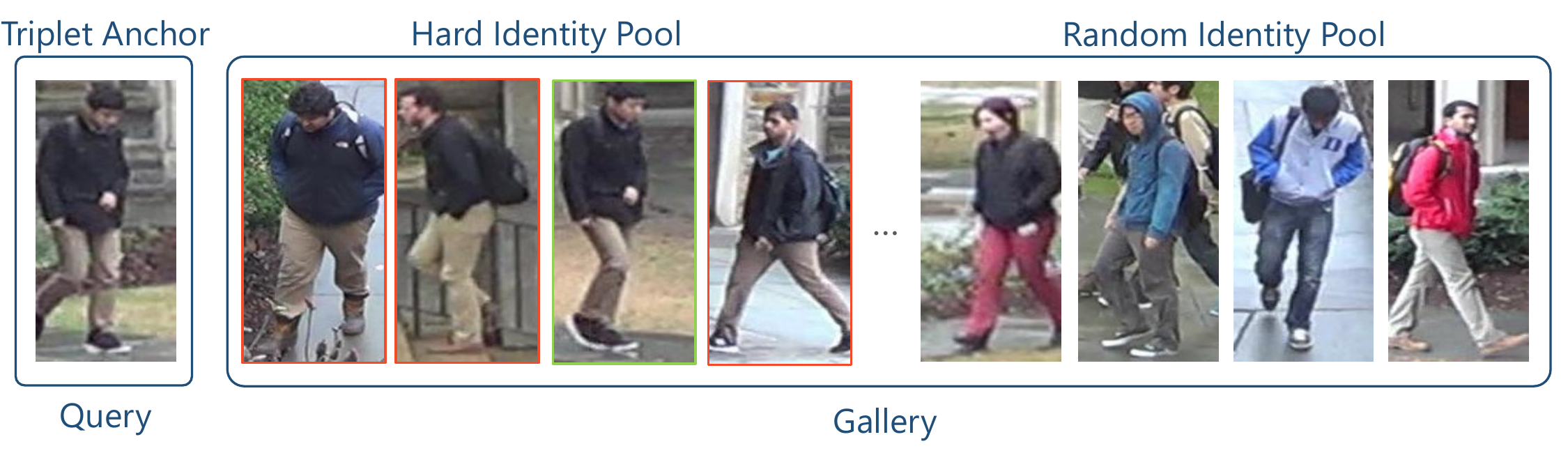}
    	\caption{Hard identity mining: For each anchor identity, half of the $P-1$ identities in the batch are sampled from the hard identity pool, the other half from the random identity pool. Hard-negative identities (correct matches) are outlined in red (green).}
        \label{fig:hard_mining}
    \end{minipage}
\end{figure*}

\subsection{Learning Appearance Features} 

Given a large collection of labeled person snapshots we learn appearance features using an adaptive weighted triplet loss. For an anchor sample $x_a$, positive samples $x_p \in P(a)$ and negative samples $x_n \in N(a)$, we re-write the triplet loss in its most general form as:

\begin{equation}
L_{3} = \left[ m + \sum_{\mathclap{x_p \in P(a)}} \hspace{1pt} w_p d(x_a,x_p) - \sum_{\mathclap{x_n \in N(a)}} \hspace{1pt} w_n d(x_a,x_n) \right]_+
\label{eq:triplet_general}
\end{equation}

where $m$ is the given inter-person separation margin, $d$ denotes distance of appearance, and $[\boldsymbol{\cdot}]_+ = max(0,\boldsymbol{\cdot})$. This reformulation has two advantages. First it avoids the combinatorial process of triplet generation by using all the samples rather than a selection. Instead, the challenge of learning good features is to assign larger weights to difficult positive and negative samples. Second, the positive/negative class imbalance is easily handled by reflecting it in the weight distribution.

Hermans \emph{et al}~\cite{hermans2017defense} and Mischuk \emph{et al}~\cite{mishchuk2017working} have proposed the batch-hard triplet loss with built-in hard sample mining. The batch-hard loss weights for Equation \ref{eq:triplet_general} are binary in their approach, as the loss considers only the most difficult positive and negative sample:

\begin{equation}
w_p = \Big[x_p == \arg\max\limits_{\substack{x \in P(a)}} d(x_a, x) \Big]
\label{eq:pos_weight_hard}
\end{equation}

\begin{equation}
w_n = \Big[x_n == \arg\min\limits_{\substack{x \in N(a)}} d(x_a, x) \Big]
\label{eq:neg_weight_hard}
\end{equation}
where $[\boldsymbol{\cdot}]$ denotes the Iverson bracket. This loss gives better results than the original triplet loss with uniform weights~\cite{schroff2015facenet} because the latter washes out the contribution of hard samples and is driven to worse local minima by easy samples. On the other hand, the uniformly weighted loss is more robust to outliers because they cannot affect the weights. 

Can we define weights such that $L_{3}$ converges to parameters at least as good as the batch-hard loss, yet remains robust to outliers? Our first improvement pertains to weights that achieve high accuracy and training stability \emph{simultaneously}.  Equations \ref{eq:pos_weight_hard}-\ref{eq:neg_weight_hard} assign full weight to the hardest positive/negative sample for each anchor while ignoring the remaining positive and negative samples. Instead, we assign adaptive weights using the softmax/min weight distributions as follows (see Figure \ref{fig:weighted_loss}):

\begin{equation}
w_p = \frac{ e^{d(x_a, x_p)} }{ \sum\limits_{\substack{x \in P(a)}} e^{d(x_a, x)} } \ , \quad w_n = \frac{ e^{-d(x_a, x_n)} }{ \sum\limits_{\substack{x \in N(a)}} e^{-d(x_a, x)} } \ .
\label{eq:adaptive_weights}
\end{equation}

The adaptive weights in Equation \ref{eq:adaptive_weights} give little importance to easy samples and emphasize the most difficult ones. When several difficult samples appear in a batch, they all get their fair share of the weight. This differs from the hard weight assignments of Equations \ref{eq:pos_weight_hard}-\ref{eq:neg_weight_hard} which give importance to the \emph{single} most difficult sample. Adaptive weights are useful when the most difficult sample in a batch is an outlier, yet there exist other difficult samples to learn from. Experiments in such cases demonstrate the favorable properties of adaptive weights.

For batch construction during training we leverage the idea of $PK$ batches also introduced by Hermans \emph{et al}~\cite{hermans2017defense}. In each batch there are $K$ sample images for each of $P$ identities. This approach has shown very good performance in similarity-based ranking and avoids the need to generate a combinatorial number of triplets. During a training epoch each identity is selected in its batch in turn, and the remaining $P-1$ batch identities are sampled at random. $K$ samples for each identity are then also selected at random.

Our second improvement is on the procedure that selects difficult identities. As the size of the training set increases, sampling $P-1$ identities at random rarely picks the hardest negatives, thereby moderating batch difficulty. This effect can also be observed in the last few epochs of training, when many triplets within a batch exhibit zero loss. 

To increase the chances of seeing hard negatives, we construct two sets to sample identities from. An example is shown in Figure~\ref{fig:hard_mining}. The hard identity pool consists of the $H$ most difficult identities given the anchor, and the random identity pool consists of the remaining identities. Then in a $PK$ batch of an anchor identity we sample the remaining $P-1$ identities from the hard or random identity pool with equal probability. This technique samples hard negatives more frequently and yet the batch partially preserves dataset statistics by drawing random identities. The pools can be constructed either after training the network for few epochs, or computed from a pre-trained network. We demonstrate the benefit of using the hard-identity mining scheme in the experiments section. 

%The hardest positive/negative samples still dominate the loss, while all other samples of a $PK$ batch contribute. This modification makes training stable and improves in rank accuracy.

\subsection{MTMC Tracker}

Given an array $O_d$ of $k$-dimensional detections as input, the tracker outputs a $(k+1) \times o_t$ array $O_t = f_t(\Theta_t, O_d)$ of $o_t$ detections. The added dimension is the identity label assignment to the input observations. In our design, the tracker first computes features for all $o_d$ input observations, then estimates correlation between all pairs of features, and finally solves a correlation clustering problem to assign identities to observations. Two post-processing steps, interpolation and pruning, interpolate detections to fill gaps and remove trajectories with low confidence. For this reason, the number $o_t$ of output detections can differ from the number $o_d$ of input detections.

\noindent\textbf{Detector.} We use the off-the-shelf OpenPose person detector which achieves good performance~\cite{cao2017realtime}. This detector learns part affinity fields to capture the relation between body parts and applies greedy parsing to combine part affinities into bounding boxes. During training it is supervised directly on part affinities rather than bounding box accuracy.
% A detector takes a video $V$ as input and produces as output a $k \times o_d$ array $O_d$ of $o_d$ detections with $k$ parameters each. Typically $k=5$ (bounding box and a timestamp). More sophisticated detectors may also output detection confidence or pose information.

\noindent\textbf{Appearance Features.} We use the ResNet50 model pre-trained on ImageNet and follow its \textit{pool5} layer by a dense layer with 1024 units, batch normalization, and ReLU. Another dense layer yields 128-dimensional appearance features. We train the model with the adaptive weighted triplet loss, data augmentation, and hard-identity mining.

We define the appearance correlation between two detections as $w_{ij} = \frac{t_a - d(x_i,x_j)}{t_a}$ (a number $\leq 1$) where the threshold $t_a = \frac{1}{2}(\mu_p + \mu_n)$ separates the means of positive and negative distances $\mu_p$ and $\mu_n$ of all training pairs.

\noindent\textbf{Data Augmentation.} We augment the training images online with crops and horizontal flips to compensate for detector localization errors and to gain some degree of viewpoint/pose invariance. For illumination invariance we additionally apply contrast normalization, grayscale and color multiplication effects on the image. For resolution invariance we apply Gaussian blur of varying $\sigma$. For additional viewpoint/pose invariance we apply perspective transformations and small distortions. We additionally hide small rectangular image patches to simulate occlusion.

\noindent\textbf{Motion Correlation.} We use a linear motion model to predict motion correlation. As the forward-backward prediction error $e_m = e_f + e_b$ is non-negative, we use the trajectories from the training set to learn a threshold $t_m$ that separates positive and negative evidence, and a scaling factor $\alpha$ to convert errors to correlations: $w_m = \alpha (t_m - e_m)$. Impossible associations receive correlation $w_m = -\infty$.

\noindent\textbf{Optimization.} A matrix $W = (W_a + W_m)\odot D$ collects appearance and motion correlations, and the matrix $D$ specifies discounts $ = e^{-\beta \Delta t}\in [0,1]$ that decay correlations to zero as the time lag $\Delta t$ between observation increases. $D$ ensures association of time-distant trajectories only if there is a chain of associations with positive net correlation that connects them. Parameters $t_m$, $\alpha$, $\beta$ are chosen to maximize tracking accuracy over small subsets of the training set.

We establish co-identity by correlation clustering. Given a weighted graph $G = (V,E,W)$, two nodes $v_i$ and $v_j$ are co-identical if the binary incidence variable $x_{ij}=1$ in the solution. Correlation clustering is defined as:

\begin{equation}
X^* = \arg\max\limits_{\{x_{ij}\}} \sum_{(i,j) \in E} w_{ij} x_{ij}
\label{eq:CC}
\end{equation}
\begin{equation}
	\text{subject to:} \quad x_{ij} + x_{jk} \leq 1 + x_{ik} \quad \forall i,j,k \in V\label{eq:triangular_constraint}
\end{equation}

Equation \ref{eq:CC} maximizes positive (negative) correlation within (between) clusters and the constraints in Equation \ref{eq:triangular_constraint} enforce transitivity in the solution.

\noindent\textbf{Multi-Level Reasoning.} Our method reduces the computational burden by reasoning hierarchically over three levels. The first level computes one-second long tracklets, the second associates tracklets into single-camera trajectories, and the third associates single-camera trajectories into multi-camera identities.  

Tracklets are found in disjoint, one-second long windows. Trajectories are computed online in a sliding temporal window that overlaps 50\% with the previous window. All trajectories that have at least one detection in the window are re-considered for association. We set the window width for single-camera trajectories to 10 seconds, and 1.5 minutes for multi-camera trajectories.

\section{Experiments}

We run the following experiments on recent benchmarks for MTMCT and Re-ID: (a) Measure overall MTMCT performance, (b) measure the impact of improved detector and features during tracking, (c) study the relation between measures of accuracy for ranking and tracking, (d) demonstrate the usefulness of the adaptive weighted triplet loss and hard negative mining, and (e) analyze tracker failures. 

\subsection{Benchmarks}

\noindent\textbf{DukeMTMC}~\cite{ristani2016performance} is a large-scale tracking dataset recorded on the Duke University campus featuring 2.8k identities, of which 1.8k belong to the training/validation set. The dataset was recorded by 8 cameras with 1080p 60fps image quality and the evaluation is done on disjoint fields of view. The video duration of each camera is 1 hour and 25 minutes. We benchmark our method on the 25 minute long \textit{test-easy} sequence and 15 minute long \textit{test-hard} sequence hosted on MOTChallenge~\cite{MOTChallenge2015}. \textit{test-hard} features a group of ~50 people traveling through 4 cameras. We use the 17 minute long validation sequence for ablation experiments.

\noindent\textbf{DukeMTMC-reID}~\cite{ristani2016performance, zheng2017unlabeled} is a subset of the DukeMTMC tracking dataset~\cite{ristani2016performance} for image-based person re-identification. It features 1,404 identities appearing in more than two cameras and 408 identities who appear in only one camera are used as distractors. 702 identities are reserved for training and 702 for testing.  

\noindent\textbf{Market-1501}~\cite{zheng2015scalable} is a large-scale person re-identification dataset with 1,501 identities observed by 6 near-synchronized cameras. The dataset was collected in the campus of Tsinghua University. It features 32,668 bounding boxes obtained using the deformable parts model detector. The dataset is challenging as the boxes are often misaligned and viewpoints can differ significantly. 751 identities are reserved for training and the remaining 750 for testing. 

%\paragraph{MARS~\cite{zheng2016mars}} is a large-scale video-based person re-identification dataset extending Market-1501~\cite{zheng2015scalable}. The dataset was recorded by 6 cameras, 5 with 1080p quality and 1 with SD quality. MARS consists of 1,261 different pedestrians captured by at least 2 cameras. It features 20,478 tracklets and 1,191,003 bounding boxes. 625 identities are reserved for training and the remaining 636 for testing.

\begin{table*}[h]
	\small
	\begin{center}
		\begin{tabular}{ |l | c | c | c ? c | c | c ? c | c | c | c ? c | c | c | c |  }
			\hline
			& \multicolumn{3}{|c ?}{Multi-Camera Easy} & \multicolumn{3}{|c ?}{Multi-Camera Hard} & \multicolumn{4}{|c ?}{Single-Camera Easy} & \multicolumn{4}{|c |}{Single-Camera Hard} \\ \hline
			
 & IDF1  & IDP  & IDR & IDF1  & IDP  & IDR & IDF1  & IDP  & IDR & MOTA & IDF1  & IDP  & IDR & MOTA\\ \hline
			
			BIPCC~\cite{ristani2016performance}   &56.2  &67.0  &48.4   &47.3 &59.6& 39.2  & 70.1 &	83.6 &	60.4 & 59.4  & 64.5 &	81.2&	53.5&	54.6\\ \hline
			lx\_b~\cite{liang2017multi}					  & 58.0 & 72.6 & 48.2  &48.3 & 60.6 & 40.2& 70.3 &	88.1 & 58.5 &	61.3 & 64.2	&80.4&	53.4&	53.6\\ \hline
			PT\_BIPCC~\cite{maksai2017} & - & - & - & - & - & - & 71.2 &	84.8 &	61.4 &	59.3&  65.0 &	81.8 &	54.0 & 54.4 \\ \hline
			MTMC\_CDSC~\cite{tesfaye2017multi}    & 60.0 & 68.3 & 53.5  &50.9 & 63.2 & 42.6& 77.0 &	87.6 & 68.6	&	70.9 & 65.5&	81.4	&54.7&	59.6\\ \hline
			MYTRACKER~\cite{yoon2018mht}  & 64.8 &	70.8 &	59.8 & 47.3 &	55.6 &	41.2 & 80.0 &	87.5 &	73.8 & 77.7 & 63.4 &	74.5 &	55.2 &	59.0 \\ \hline
%			\multicolumn{15}{|c|}{Private Detector} \\ \hline
			
			MTMC\_ReID~\cite{zhang2017multi}$^\dagger$  				  & 78.3 & 82.6 & 74.3  & 67.7 & \textbf{78.6} & 59.4 &  86.3&	91.2&	82.0&	83.6 & 77.6&	\textbf{90.1} &	68.1&	69.6\\ \hline
			\hline
			\textbf{DeepCC}       				  & \textbf{82.0} & \textbf{84.3} & \textbf{79.8} & \textbf{68.5} &  75.8 &  \textbf{62.4} &  \textbf{89.2 } & \textbf{91.7} & \textbf{86.7} & \textbf{87.5}  & \textbf{79.0} &87.4  & \textbf{72.0}& \textbf{70.0} \\ \hline
			
		\end{tabular}
	\end{center}
	\caption{DukeMTMCT results. Methods in $^\dagger$ are unrefereed submissions. }
	
	\label{tab:duke_results}
\end{table*}

\subsection{Evaluation}

For MTMCT evaluation we use ID measures of performance~\cite{ristani2016performance} which indicate how well a tracker identifies who is where regardless of  where or why mistakes occur. IDP (IDR) is the fraction of computed (true) detections that are correctly identified. IDF1 is the ratio of correctly identified detections over the average number of true and computed detections. IDF1 is used as the principal measure for ranking MTMC trackers. ID measures first compute a 1-1 mapping between true and computed identities that maximizes true positives, and then compute the ID scores.

For single-camera evaluation we also report MOTA, which counts mistakes by how often, not how long, incorrect decisions are made. MOTA is based on the CLEAR-MOT mapping~\cite{bernardin08} which under-reports multi-camera errors, therefore we report it only in single camera experiments.

For person re-identification experiments we report rank accuracy as well as mean average precision (mAP)~\cite{zheng2015scalable}.

\subsection{Model Training}

For training we set $P=18$, $K=4$, $m=1$, resolution 256$\times$128. The learning rate is $3\cdot10^{-4}$ for the first 15000 iterations, and decays to $10^{-7}$ at iteration 25000. In experiments with hard identity mining we construct the hard and random pools once with features obtained at iteration 5000, then sample identities from these pools until the last iteration. The hard identity pool size $H$ is set to 50 and we found that similar scores were obtained with 30-100 identities (4\%-15\% of all training identities). Extreme sizes yield little gain: A size of 1 contains a single hard identity which can be an outlier, a large HN pool nears random sampling.

\section{Results}

We discuss results for MTMC tracking, where our proposed method outperforms previous and concurrent work in IDF1 score and identity recall IDR; study the influence of different components; and analyze common tracking failures. We also present results on person re-identification datasets, where our learned appearance features achieve competitive results.

\subsection{Impact of Learning}

We evaluate how detector and feature choice impact multi-camera IDF1 on the DukeMTMC validation set. Results are shown in Table \ref{tab:ablation}.

\begin{table}[H]
	\small
	\begin{center}
		\begin{tabular}{ |l | c | c | c |}
			\hline
			& IDF1  & IDP  & IDR \\ \hline
			BIPCC  (DPM + HSV)~\cite{ristani2016performance}       & 54.98  & 62.67  & 48.97 \\ 
			DeepCC (OpenPose + HSV) & 58.24  & 60.60  & 56.06 \\ 
			DeepCC (DPM + ResNet) & 65.68  & 74.87  & 58.50 \\ 
			\hline
			\textbf{DeepCC} (OpenPose + ResNet) & \textbf{80.26} & \textbf{83.50} & \textbf{77.25} \\ \hline
			
		\end{tabular}
	\end{center}
	\caption{Impact of improving detector and features on multi-camera  performance for the validation sequence.}
	\label{tab:ablation}
\end{table}

First we compare the behavior of our baseline method BIPCC with and without deep features. BIPCC uses part based color histograms as appearance features. Our learned features play an important role in improving IDF1 by 10.7 points (third row) in multi-camera performance. 

Second we measure the impact of the deep learned detector. We substituted the baseline's DPM detections (first row) with those obtained from OpenPose~\cite{cao2017realtime} (second row). Although  single-camera IDF1 increases from 75.0 to 85.5, multi-camera IDF1 increases by only 3.26 points (from 54.98 to 58.24\%). This indicates that the detector plays an important role in single-camera tracking by reducing false negatives, but in multi-camera tracking weak features take little advantage of better single-camera trajectories. 

These results imply that good features are crucial for MTMC tracking, and that a good detector is most useful for improving single-camera performance. The best MTMCT performance is achieved by combining both.

\begin{table*}[h]
	\tiny
	\begin{center}
		{\setlength{\tabcolsep}{1em}%  1 is the default, change whatever you need
		\begin{tabular}{ |c|c|c|c|c|c|c?c|c|c|c|c|c?c|c|c|c|c|c?c|c|c|c|c|c|  }
			\hline
		    & \rotatebox[origin=l]{90}{BIPCC~\cite{ristani2016performance}} & \rotatebox[origin=l]{90}{PT\_BIPCC~\cite{maksai2017}} & \rotatebox[origin=l]{90}{MTMC\_CDSC~\cite{tesfaye2017multi}} & \rotatebox[origin=l]{90}{MYTRACKER~\cite{yoon2018mht}} & \rotatebox[origin=l]{90}{MTMC\_ReID~\cite{zhang2017multi}$^\dagger$} & \rotatebox[origin=l]{90}{\textbf{DeepCC}} 
		    & \rotatebox[origin=l]{90}{BIPCC~\cite{ristani2016performance}} & \rotatebox[origin=l]{90}{PT\_BIPCC~\cite{maksai2017}} & \rotatebox[origin=l]{90}{MTMC\_CDSC~\cite{tesfaye2017multi}} & \rotatebox[origin=l]{90}{MYTRACKER~\cite{yoon2018mht}} & \rotatebox[origin=l]{90}{MTMC\_ReID~\cite{zhang2017multi}$^\dagger$} & \rotatebox[origin=l]{90}{\textbf{DeepCC}} 
		    & \rotatebox[origin=l]{90}{BIPCC~\cite{ristani2016performance}} & \rotatebox[origin=l]{90}{PT\_BIPCC~\cite{maksai2017}} & \rotatebox[origin=l]{90}{MTMC\_CDSC~\cite{tesfaye2017multi}} & \rotatebox[origin=l]{90}{MYTRACKER~\cite{yoon2018mht}} & \rotatebox[origin=l]{90}{MTMC\_ReID~\cite{zhang2017multi}$^\dagger$} & \rotatebox[origin=l]{90}{\textbf{DeepCC}} 
		    & \rotatebox[origin=l]{90}{BIPCC~\cite{ristani2016performance}} & \rotatebox[origin=l]{90}{PT\_BIPCC~\cite{maksai2017}} & \rotatebox[origin=l]{90}{MTMC\_CDSC~\cite{tesfaye2017multi}} & \rotatebox[origin=l]{90}{MYTRACKER~\cite{yoon2018mht}} & \rotatebox[origin=l]{90}{MTMC\_ReID~\cite{zhang2017multi}$^\dagger$} & \rotatebox[origin=l]{90}{\textbf{DeepCC}}

		     \\ \hline
			& \multicolumn{6}{c?}{MOTA} &  \multicolumn{6}{c?}{IDP} & \multicolumn{6}{c?}{IDR} & \multicolumn{6}{c|}{IDF1}\\ \hline
			
						Easy-all  & 59.4	& 59.3 & 70.9 & 77.7 & 83.6	& \textbf{87.5} & 83.6 & 84.8 & 87.6 & 87.5	& 91.2 & \textbf{91.7} & 60.4 & 61.4 & 68.6 & 73.8 & 82.0 & \textbf{86.7}  & 70.1 & 71.2 & 77.0 & 80.0 & 86.3 & \textbf{89.2} \\ \hline 
			
			Cam1 & 43.0 & 42.9 & 69.9 & 84.9 & 87.4 & \textbf{93.3} & 91.2 & 91.9 & 89.1 & 89.7 & 91.1 & \textbf{95.6} & 41.8 & 42.2 & 67.7 & 79.6 & 86.2 & \textbf{93.0} & 57.3 & 57.8 & 76.9 & 84.3 & 88.6 & \textbf{94.3}\\ 
			Cam2 & 44.8 & 44.7 & 71.5 & 78.4 & 84.2 & \textbf{87.1} & 69.3 & 70.4 & 90.9 & 88.9 & 92.4 & \textbf{93.6} & 67.1 & 68.0 & 73.4 & 75.9 & 82.9 & \textbf{87.4} & 68.2 & 69.2 & 81.2 & 81.9 & 87.4 & \textbf{90.4}\\ 
			Cam3 & 57.8 & 57.8 & 67.4 & 65.7 & \textbf{82.4} & 79.7 & 78.9 & 78.2 & 76.3 & 76.2 & \textbf{87.8} & 86.2 & 48.8 & 48.4 & 56.0 & 63.5 & \textbf{79.7} & 77.7 & 60.3 & 59.8 & 64.6 & 69.3 & \textbf{83.6} & 81.8\\ 
			Cam4 & 63.2 & 63.2 & 76.8 & 79.8 & \textbf{91.9} & 91.8 & 88.7 & 91.7 & 91.2 & 84.1 & \textbf{97.7} & 96.3 & 62.8 & 64.9 & 79.0 & 77.6 & 93.1 & \textbf{94.4} & 73.5 & 76.0 & 84.7 & 80.7 & \textbf{95.4} & 95.3\\ 
			Cam5 & 72.8 & 72.6 & 68.9 & 76.6 & 80.8 & \textbf{86.2} & 83.0 & 83.0 & 76.1 & 81.4 & \textbf{87.2} & 83.6 & 65.4 & 65.6 & 61.9 & 67.3 & 75.8 & \textbf{77.7} & 73.2 & 73.3 & 68.3 & 73.7 & \textbf{81.1} & 80.6\\ 
			Cam6 & 73.4 & 73.4 & 77.0 & 82.8 & 83.1 & \textbf{88.7} & 87.5 & 91.7 & 91.6 & 88.9 & 91.7 & \textbf{93.4} & 69.1 & 72.4 & 75.3 & 78.8 & 82.5 & \textbf{92.2} & 77.2 & 80.9 & 82.7 & 83.5 & 86.9 & \textbf{92.8}\\ 
			Cam7 & 71.4 & 71.4 & 73.8 & 77.0 & 80.8 & \textbf{82.2} & 93.6 & 93.6 & \textbf{94.0} & 91.4 & 92.8 & 93.7 & 70.6 & 70.6 & 72.5 & 73.5 & 80.1 & \textbf{83.7} & 80.5 & 80.5 & 81.8 & 81.5 & 86.0 & \textbf{88.5}\\ 
			Cam8 & 60.7 & 60.9 & 63.4 & 71.6 & 79.9 & \textbf{85.0} & \textbf{92.2} & \textbf{92.2} & 89.1 & 90.8 & 91.1 & 89.4 & 59.6 & 60.0 & 61.8 & 71.3 & 78.6 & \textbf{82.4} & 72.4 & 72.7 & 73.0 & 79.9 & 84.4 & \textbf{85.8}\\ \hline \hline

			Hard-all & 54.6 & 54.4 & 59.6 & 59.0 & 69.6 & \textbf{70.0} & 81.2 & 81.8 & 81.4 & 74.5 & \textbf{90.1} & 87.4 &  53.5 & 54.0 & 54.7 & 55.2 & 68.1 & \textbf{72.0} & 64.5 & 65.0 & 65.5 & 63.4 & 77.6 & \textbf{79.0}
			  \\ \hline
			Cam1 & 37.8 & 37.4 & 63.2 & 61.1 & 74.4 & \textbf{79.6} & 92.5 & 91.9 & 83.0 & 72.2 & 92.3 & \textbf{94.7} & 36.8 & 36.7 & 56.4 & 58.4 & 76.1 & \textbf{80.1} & 52.7 & 52.5 & 67.1 & 64.6 & 83.4 & \textbf{86.8}\\ 
			Cam2 & 47.3 & 46.6 & 54.8 & 50.4 & \textbf{70.9} & 57.9 & 65.7 & 66.0 & 78.8 & 61.2 & \textbf{89.1} & 77.5 & 56.1 & 56.7 & 53.1 & 52.6 & 66.7 & \textbf{67.3} & 60.6 & 61.0 & 63.4 & 56.6 & \textbf{76.3} & 72.0\\ 
			Cam3 & 46.7 & 46.7 & 68.8 & 70.3 & \textbf{87.1} & 84.2 & \textbf{96.1} & 96.1 & 91.1 & 86.9 & 94.9 & 90.8 & 46.5 & 46.5 & 73.7 & 74.1 & \textbf{89.2} & 87.1 & 62.7 & 62.7 & 81.5 & 80.0 & \textbf{91.9} & 88.9\\ 
			Cam4 & 85.3 & 85.5 & 75.6 & 81.2 & \textbf{95.0} & 90.3 & 86.0 & 93.6 & 87.1 & 84.4 & \textbf{97.3} & 93.0 & 82.7 & 91.0 & 78.1 & 82.2 & \textbf{97.7} & 97.0 & 84.3 & 92.3 & 82.3 & 83.3 & \textbf{97.5} & 94.9\\ 
			Cam5 & 78.3 & 78.3 & 78.6 & 81.9 & 77.2 & \textbf{86.0} & 90.1 & 90.1 & 91.5 & \textbf{93.3} & 88.4 & 90.9 & 75.1 & 75.1 & 75.7 & 79.2 & 75.3 & \textbf{85.5} & 81.9 & 81.9 & 82.8 & 85.7 & 81.3 & \textbf{88.1}\\ 
			Cam6 & 59.4 & 59.4 & 53.3 & 56.1 & 58.4 & \textbf{63.3} & 81.7 & 82.4 & 71.2 & 70.0 & 86.3 & \textbf{87.0} & 52.7 & 53.3 & 42.3 & 44.9 & 55.4 & \textbf{62.2} & 64.1 & 64.7 & 53.1 & 54.7 & 67.5 & \textbf{72.5}\\ 
			Cam7 & 50.8 & 50.6 & 50.8 & 49.8 & 60.3 & \textbf{61.4} & 81.2 & 81.4 & 84.7 & 74.7 & \textbf{91.4} & 85.2 & 47.1 & 47.2 & 47.1 & 44.4 & 59.7 & \textbf{61.3} & 59.6 & 59.8 & 60.6 & 55.7 & \textbf{72.2} & 71.3\\ 
			Cam8 & 73.0 & 73.0 & 70.0 & 71.5 & \textbf{85.6} & 85.0 & \textbf{94.9} & 94.9 & 90.3 & 93.5 & 92.2 & 92.3 & 72.8 & 72.8 & 73.9 & 70.5 & 83.7 & \textbf{87.7} & 82.4 & 82.4 & 81.3 & 80.4 & 87.7 & \textbf{89.9}\\ \hline
		\end{tabular}
	
	}
	\end{center}
	\caption{Detailed DukeMTMCT single-camera tracking results for the \textit{test-easy} and \textit{test-hard} sequences. Methods in $^\dagger$ are unrefereed submissions.}
	\label{tab:results_sc} 
	\end{table*}

\subsection{MTMC Tracking}

Overall results are presented in Tables \ref{tab:duke_results} and \ref{tab:results_sc}. Our method DeepCC improves the multi-camera IDF1 accuracy w.r.t to the previous state of the art MTMC\_CDSC~\cite{tesfaye2017multi} by 22 and 17.6 points for the \textit{test-easy} and \textit{test-hard} sequences, respectively. For the single-camera easy and hard sequences, the IDF1 improvement is 12.2 and 13.5 points, and MOTA improves by 16.6 and 10.4 points. 

Compared to unrefereed submissions, we perform slightly worse on IDP on the hard sequence. This could be due to a choice of detector that works better for crowded scenarios, a detector that is more conservative, and/or more conservative association. We nonetheless outperform all methods on IDF1, IDR and MOTA.

It is worth noting that our method achieves the highest identity recall IDR on all scenarios, and on nearly all single-camera sequences. Identity recall is Achille's heel for modern multi-target trackers, as they commonly fail to re-identify targets after occlusions~\cite{leal2017tracking}. We believe that this improvement is a combination of better detections, joint optimization, and a discriminative feature embedding.

\subsection{Impact of Loss and Hard Negative Mining}

Our Re-ID results for similarity-based ranking are shown in Tables \ref{tab:dukemtmcreid} and \ref{tab:market}. Scores are averages of five repetitions and no test-time augmentation is used. (a) Our Adaptive Weighted Triplet Loss (AWTL) consistently improves over the batch-hard loss~\cite{hermans2017defense, mishchuk2017working}. (b) When training with square Euclidean distance to emphasize sensitivity to outliers our loss is robust in all scenarios, whereas the batch-hard loss shows to be unstable on the Duke dataset. (c) The proposed hard identity mining scheme (HNM) is also beneficial, and our adaptive weighted loss is both accurate and stable with difficult batches. (d) We also compare against a recent method that combines two network streams for better performance~\cite{chen2017person}. When employing a similar technique (2-stream ensemble) we improve our ranking accuracy further.  %Yet we have shown that excess performance in re-identification benchmarks has diminishing returns in tracking.

\subsection{Accuracy of Tracking vs. Ranking}

\begin{figure}[t]
	\centering
	\includegraphics[scale = 0.45]{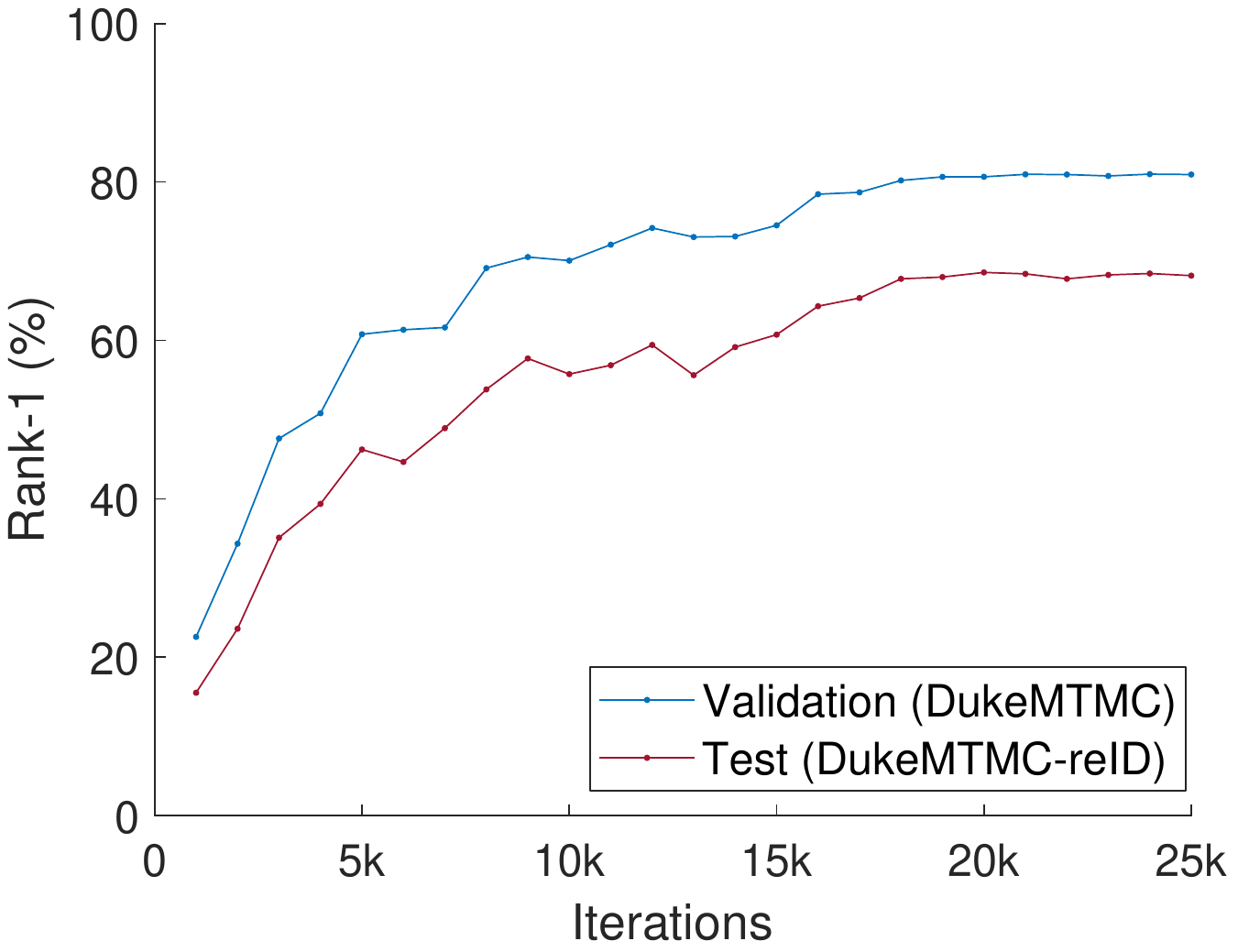}
	\caption{Relation between validation and test sets.}
	\label{fig:rank_vs_train}
\end{figure}

\begin{figure}[t]
	\centering
	\includegraphics[scale = 0.45]{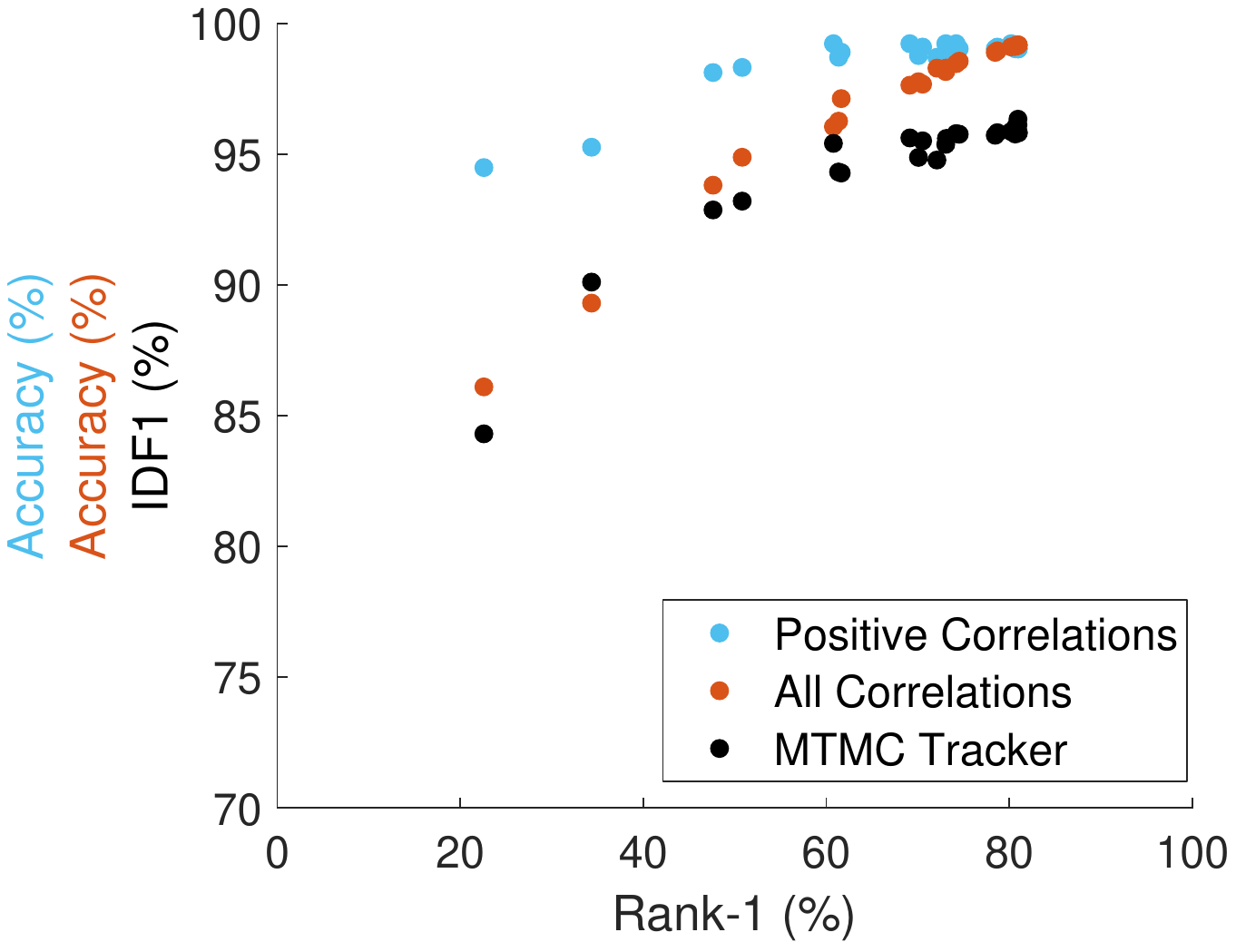}
	\caption{Relation of tracking, correlation, and rank accuracy.}
	\label{fig:tracking_vs_rank}
\end{figure}

As more and more re-identification methods are being applied to multi-target tracking, we study the relation between ID measures for MTMC tracking and rank measures for Re-ID. In this experiment, we freeze ground truth single-camera trajectories and perform across-camera tracking with features at different times during training, resulting in different levels of ranking accuracy. Appearance features are learned from scratch using the 461 DukeMTMC-reID training IDs that do not appear in the validation sequence. Tracking accuracy is evaluated on the DukeMTMC validation sequence (241 IDs), and rank-1 accuracy on both DukeMTMC-reID test (702 IDs) and DukeMTMC validation. Results are shown in Figures \ref{fig:rank_vs_train}-\ref{fig:tracking_vs_rank}.

We observe the following: Figure \ref{fig:rank_vs_train}: Rank-1 accuracy for DukeMTMC-reID test and DukeMTMC validation correlate, even if the former is more difficult than the latter due to 3x as many identities. Figure \ref{fig:tracking_vs_rank}: (a) Features with modest rank-1 performance can still do well in MTMCT because of more limited and diverse identities to compare between, and because tracking is also helped by motion information. (b) MTMCT IDF1 performance improves with rank-1 accuracy. However, after a point, further improvement in rank-1 accuracy yields diminishing returns in IDF1.

Our interpretation for this saturation effect is as follows. Initially, the Re-ID model learns to separate positive and negative samples, and tracking performance increases linearly with rank-1 performance. Once enough correlations have the correct sign, correlation clustering can infer the remaining missing agreements by enforcing transitivity (inequality \ref{eq:triangular_constraint}). Therefore, correcting the sign of the remaining correlations has a smaller effect on IDF1. Even beyond that point, the Re-ID model tries to satisfy the separation margin of $L_3$ by further pulling co-identical samples together and non co-identical ones apart. These changes do not affect the correlation signs and have little influence on IDF1.

\begin{table}[t]
	\footnotesize
	\begin{center}
		\begin{tabular}{ |l | c  c | c  c|}
			\hline
			& \multicolumn{2}{c|}{Euclidean}  & \multicolumn{2}{c|}{SqEuclidean}    \\ 
			& mAP & rank-1 & mAP & rank-1   \\ \hline
			BoW+KISSME~\cite{zheng2015scalable}  &  12.17 & 25.13 & - & -  \\ 
			LOMO+XQDA~\cite{liao2015person} & 17.04 & 30.75 & - & -\\ 
			Baseline~\cite{zheng2016person} 	& 44.99 & 65.22 & - & -\\ 
			PAN~\cite{zheng2017pedestrian}       & 	51.51 & 71.59  & - & -\\ 
			SVDNet~\cite{SunZDW17} & 56.80 & 76.70 & - & -\\ \hline
			TriHard~\cite{hermans2017defense}  & 54.60 & 73.24 & 0.28 & 0.89   \\ 
   \textbf{AWTL}  & \textbf{54.97} & \textbf{74.23} & \textbf{52.37} & \textbf{71.45}  \\ \hline
   
   TriHard  (+Aug) & 56.65 & 74.91 & 0.48 & 1.25  \\ 
   \textbf{AWTL (+Aug)} & \textbf{57.28} & \textbf{75.31} & \textbf{55.94} & \textbf{75.04}  \\ \hline
			
	   TriHard (+Aug+HNM) & 54.90 & 74.23 & 0.30 & 0.94 \\ 
   \textbf{AWTL (+Aug+HNM)} & \textbf{58.74} & \textbf{77.69} & \textbf{57.84} & \textbf{76.21}  \\ \hline
   DPFL (1-stream)~\cite{chen2017person} & 48.90 & 70.10 & - & -  \\ 
      DPFL (2-stream)~\cite{chen2017person} & 60.60 & 79.20 & - & -  \\ 
  \textbf{AWTL (2-stream)} & \textbf{63.40} & \textbf{79.80} & \textbf{63.27} & \textbf{79.08} \\ \hline

		\end{tabular}
	\end{center}
	\caption{Re-ID results on DukeMTMC-ReID}
	\label{tab:dukemtmcreid}
\end{table}

\subsection{Weakness Analysis}

We analyze the one-to-one ID mapping between true and computed trajectories to understand failures in the DukeMTMC validation sequence. During evaluation, each true trajectory that is mapped to an actual computed trajectory (not a false positive) has its own ID recall, as some of its detections could be missed by the tracker. Similarly, the computed trajectories have their own precision, as they can contain false positive detections.

\begin{table}[t]
	\footnotesize
	\begin{center}
		\begin{tabular}{ |l | c  c | c c|}
			\hline
			& \multicolumn{2}{c|}{Euclidean}  & \multicolumn{2}{c|}{SqEuclidean}    \\ \hline

			& mAP & rank-1  & mAP & rank-1    \\ \hline
			DNS~\cite{zhang2016learning}  & 29.87 & 55.43 & - & -  \\ 
			GatedSiamese~\cite{varior2016gated}  & 39.55 & 65.88 & - &- \\   
			PointSet~\cite{zhou2017point} & 44.27 & 70.72 & - & -  \\ 
			SomaNet~\cite{barbosa2017looking}  & 47.89 & 73.87 & 	- & - \\ 
			PAN~\cite{zheng2017pedestrian} & 63.35 & 82.81 & - & - \\ 
             \hline

   TriHard~\cite{hermans2017defense}  & 66.63 & 82.99 & 64.47 & 82.01  \\ 
   \textbf{AWTL}  & \textbf{68.03} & \textbf{84.20} & \textbf{65.95} & \textbf{82.16} \\ \hline
   TriHard  (+Aug) & 69.57 & 85.14 & 68.92 & 84.12 \\ 
   \textbf{AWTL (+Aug)} & \textbf{70.83} & \textbf{86.11} & \textbf{69.64} & \textbf{84.71} \\ \hline
   TriHard (+Aug+HNM) & 71.13 & 86.40 & 0.16 & 0.36\\ 
   \textbf{AWTL (+Aug+HNM)} & \textbf{71.76} & \textbf{86.94} & \textbf{70.19} & \textbf{85.39} \\ \hline
   DPFL (1-stream)~\cite{chen2017person} & 66.50 & 85.70 & - & - \\ 
   DPFL (2-stream)~\cite{chen2017person} & 72.60 & 88.06 & - & - \\ 
            \textbf{AWTL (2-stream)} & \textbf{75.67} & \textbf{89.46} & \textbf{74.81} & \textbf{87.92} \\ \hline

		\end{tabular}
	\end{center}
	\caption{Re-ID results on Market-1501}
	\label{tab:market}
\end{table}

We rank computed trajectories by ID precision and true trajectories by ID recall, then inspect the trajectories with the lowest scores. This helps clarify which situations are difficult in single- and multi-camera scenarios. Single- and multi-camera scenarios are analyzed separately because their ID mapping is different.

\begin{figure}[t]
\centering
\includegraphics[height=2.1cm]{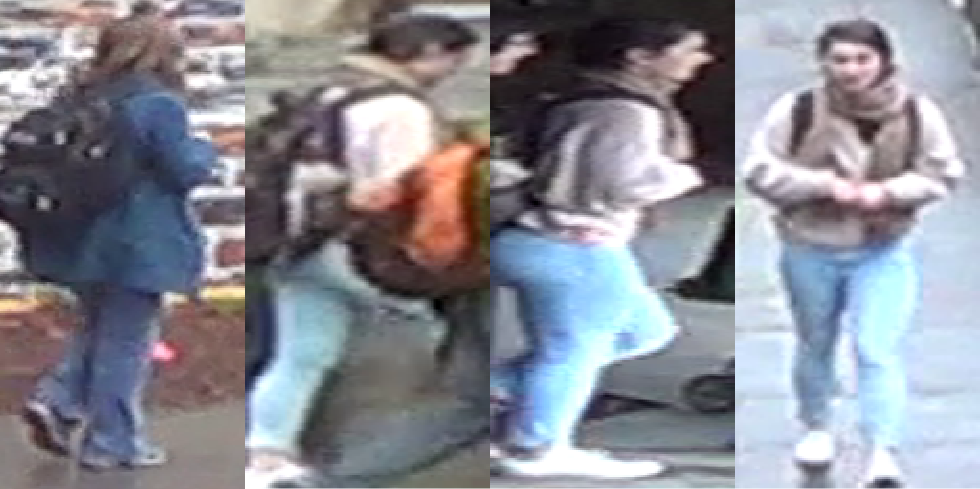} 
\includegraphics[height=2.1cm]{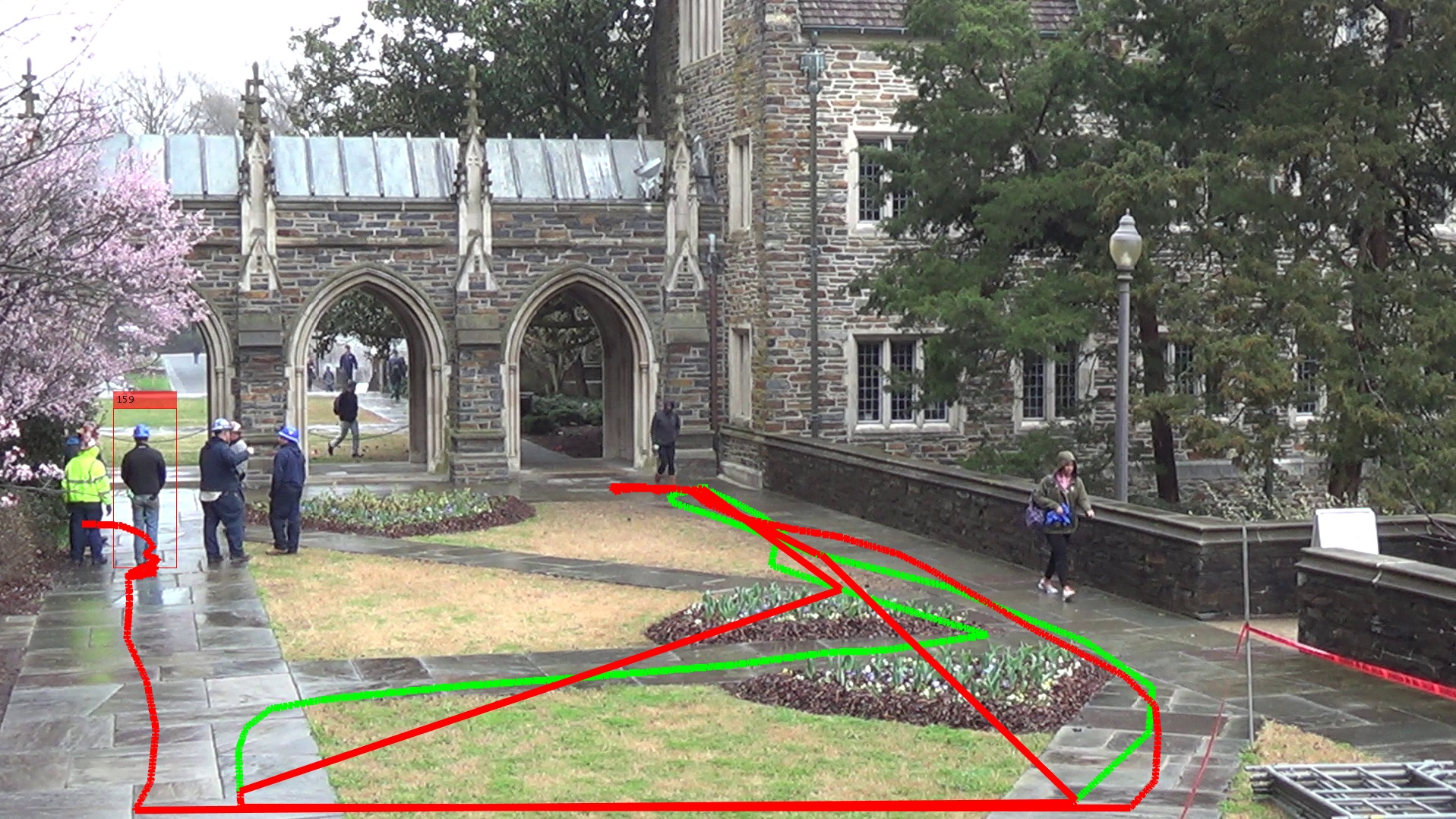}
\caption{\emph{Left}: A multi-camera trajectory with low identity precision. \emph{Right}: Example ground truth trajectory with poor identity recall in single camera tracking. Red indicates failure.}
\label{fig:mistake_sc}
\end{figure}

Two failure cases are illustrated in Figure \ref{fig:mistake_sc}. In single-camera tracking, correlations are poor when there is significant pose change, significant occlusion, and/or abrupt motion, resulting in low identity recall (left in the Figure). In multi-camera tracking, fragmentation is mostly caused by delays in blind spots and unpredictable motion. Merge errors happen in cases where people dress similarly and their inter-camera motion is plausible. 

The example in Figure \ref{fig:mistake_sc} (right) highlights one of the most difficult situations in the validation sequence, where several construction workers share similar appearance. They enter and exit the field of view a few times, and both appearance and motion correlations are weak, resulting in poor identity recall during tracking.

\section{Conclusion}

%We have introduced a method for Multi-Target Multi-Camera tracking that achieves state-of-the-art performance on the recent large-scale DukeMTMCT benchmark. The proposed tracker employs a strong detector, a discriminative appearance model, and correlation clustering optimization, and reasons hierarchically to reduce computational complexity. The proposed hard-negative mining scheme for appearance learning yields improvement in ranking accuracy on existing benchmarks. We also demonstrate that the same appearance model trained with classification loss can achieve competitive performance in tracking, even though it is inferior in ranking. We hope that other large scale MTMC tracking benchmarks will be introduced to validate our method further.

We showed that a new triplet loss with real-valued, adaptive weights, coupled with a new hard-identity mining technique that mixes difficult and random identities, yields appearance features that achieve state-of-the art performance in both MTMCT and Re-ID, whether measured by IDF1, MOTA, or rank-1 scores.

Our experiments also elucidate the relation between changes in rank-1 Re-ID score and changes in IDF1 tracking accuracy. The two performance measures relate linearly with each other at first, but the dependency saturates once rank-1 scores are good enough to yield data association correlations with the correct signs.

We hope that new large-scale data sets will be introduced to further validate our ideas.

{\small
\bibliographystyle{ieee}
\bibliography{refs}
}

\end{document}